\PassOptionsToPackage{table}{xcolor}

\documentclass[10pt,twocolumn,letterpaper]{article}

\usepackage{iccv}      

\usepackage{xcolor}

\newcommand{\ourmodel}{Opt-CWM}
\newcommand{\ourmodels}{Opt-CWM\ }
\newcommand{\psirgb}{$\boldsymbol{\Psi}^{\texttt{RGB}}$}

\usepackage{multirow} 
\usepackage{multicol} 
\usepackage{arydshln} 
\usepackage{makecell}

\usepackage{comment}
\newcommand{\highlight}[1]{\colorbox{gray!20}{\strut #1}}

\definecolor{r1c}{rgb}{0.8941,0.1097,0.1020}
\definecolor{r2c}{rgb}{0.302,0.6862,0.2902}
\definecolor{r3c}{rgb}{0.2157,0.4941,0.7216}
\definecolor{refc}{rgb}{0.21,0.49,0.74}
\definecolor{darkgolden}{rgb}{0.7,0.7,0.1}
\definecolor{darkorange}{rgb}{0.9,0.5,0.1}
\definecolor{darkred}{rgb}{0.95,0.05,0.05}

\definecolor{iccvblue}{rgb}{0.21,0.49,0.74}
\usepackage[pagebackref,breaklinks,colorlinks,allcolors=iccvblue]{hyperref}

\title{Self-Supervised Learning of Motion Concepts by Optimizing Counterfactuals}

\author{Stefan Stojanov* \quad 
David Wendt* \quad
Seungwoo Kim* \quad
Rahul Venkatesh* \quad
Kevin Feigelis \quad \\
Jiajun Wu \quad
Daniel L.K. Yamins \\
Stanford University
}

\begin{document}

\maketitle
\def\thefootnote{*}\footnotetext{Equal contribution.}\def\thefootnote{\arabic{footnote}}
\begin{abstract}
Estimating motion in videos is an essential computer vision problem with many downstream applications, including controllable video generation and robotics. Current solutions are primarily trained using synthetic data or require tuning of situation-specific heuristics, which inherently limits these models' capabilities in real-world contexts. Despite recent developments in large-scale self-supervised learning from videos, leveraging such representations for motion estimation remains relatively underexplored. 
In this work, we develop Opt-CWM, a self-supervised technique for flow and occlusion estimation from a pre-trained next-frame prediction model. Opt-CWM works by learning to optimize counterfactual probes that extract motion information from a base video model, avoiding the need for fixed heuristics while training on unrestricted video inputs. 
We achieve state-of-the-art performance for motion estimation on real-world videos while requiring no labeled data.
\footnote{Project website at: \url{https://neuroailab.github.io/opt_cwm_page/}}
\end{abstract}    
\section{Introduction}
\label{sec:intro}

The ability to extract scene motion properties such as optical flow~\cite{dosovitskiy2015flownet, teed2020raft}, occlusions, point or object tracks~\cite{harley2022particle, doersch2022tap}, and collisions or other physical events~\cite{tung2024physion, bear1physion} is important in video understanding applications such as automated video filtering~\cite{zhang2021dynamic,zhou2021event}, action recognition~\cite{kay2017kineticshumanactionvideo, soomro2012ucf101} and motion prediction~\cite{bharadhwaj2024track2act, yuangeneral}.  Recently, scene motion primitives have also been shown to be a key ingredient in improving controllability and consistency in video generation~\cite{geng2024motion} and play an important role in downstream vision applications in areas such as robotics~\cite{vecerik2024robotap, bharadhwaj2024gen2act}.  As a result, improving the estimation of motion is at the forefront of computer vision.

Optical flow and occlusion estimation are two of the core problems in this domain.  The most common approach to solving optical flow estimation uses supervised learning from labeled flow data. However, because densely annotating flow in real-world scenes is prohibitively expensive, supervised flow estimation methods usually rely on synthetic data~\cite{MIFDB16, mehl2023spring}. Methods trained on synthetic data have proven to be robust in real-world video~\cite{teed2020raft,wang2024sea}.  However, relying on this approach has limited flow estimation methods from taking advantage of recent advances in self-supervised visual representation learning from massive video datasets~\cite{tong2022videomae, qian2021spatiotemporal, feichtenhofer2021large} and inherently has to contend with a sim-to-real domain gap. 

In contrast, self-supervised flow methods are typically based on \emph{photometric loss} -- learning frame-pair feature correspondences to warp pixels from one RGB frame to corresponding locations in future frame states.  However, pure photometric loss is a weak constraint, in part because correspondences are often ill-defined (e.g., on internal portions of moving objects with homogeneous textures).
Existing state-of-the-art self-supervised motion estimation methods use various nearest neighbor or clustering procedures~\cite{jabri2020space, bian2022learning}, or complement photometric loss with strong task-specific regularizations like smoothness~\cite{jiang2024doduo, stone2021smurf}. However, because these heuristics are often only correct in narrow scenarios, the performance of such approaches is limited when the correctness conditions for the heuristic fail.  

\begin{figure*}[t!]
    \centering
    \captionsetup{type=figure}
    \includegraphics[width=\linewidth]{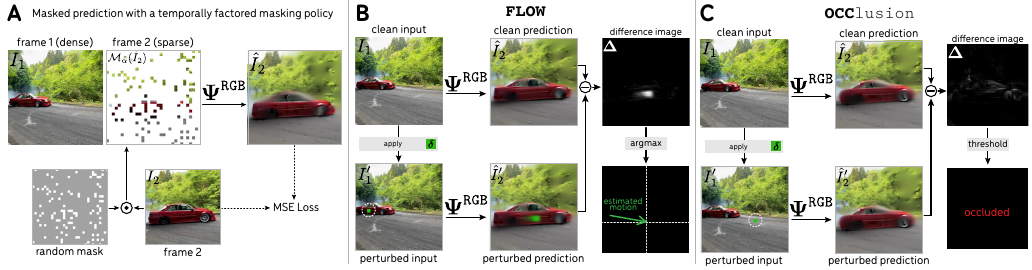}
    \vspace{-22pt}
    \caption{\textbf{Extracting flow and occlusion with counterfactual perturbation: (A)} CWMs learn to predict the next frame with a temporally factored masking policy~\cite{bear2023unifying}. \textbf{(B)} The motion of a point can be estimated using a simple counterfactual probing program \textbf{\texttt{FLOW}}: the model predicts the next frame with and without a local perturbation placed on the point, and the difference image between the clean and perturbed predictions reveals the estimated motion. \textbf{(C)} Occlusion is estimated using a related probe \textbf{\texttt{OCC}}: when the perturbation difference image is diffuse and low magnitude, that indicates the perturbed point has been occluded.}
    \label{fig:background_figure}
    \vspace{-18pt}
\end{figure*}

In this work, we investigate how to extract self-supervised flow and occlusion estimates without the use of such heuristics.  Our approach is based on Counterfactual World Modeling (CWM)~\cite{bear2023unifying, venkatesh2023understanding}, a recent method that constructs zero-shot estimates of visual properties from an underlying pre-trained multi-frame model (Figure~\ref{fig:background_figure}).
CWM begins with a \emph{sparse RGB-conditioned next frame predictor} $\boldsymbol{\Psi}^{\texttt{RGB}}$, a two-frame masked autoencoder trained with a highly asymmetric masking policy~\cite{bear2023unifying}, in which all the patches of the first frame, but very few patches of the second frame, are given to the predictor. To solve this asymmetric prediction problem, $\boldsymbol{\Psi}^{\texttt{RGB}}$ must learn to encode scene dynamics in a small number of patch feature tokens that effectively \emph{factor} temporal dynamics from visual appearance.  
Motion properties can then be extracted from the base model in a zero-shot fashion via simple ``counterfactual probes''.  For example, to compute flow from a given point in the first frame, a perturbation is made to the image at that point, and flow is computed by comparing the delta between $\boldsymbol{\Psi}^{\texttt{RGB}}$'s prediction on the perturbed (counterfactual) condition with its prediction in the original unperturbed (factual) condition (see Figure~\ref{fig:background_figure}B). Intuitively, this corresponds to placing a visual marker on the point, predictively flowing it forward, and then analyzing where it gets ``carried'' in the predicted next frame. 
The CWM approach circumvents the key limitation of the heuristic-based methods by replacing situation-specific fixed heuristics (e.g., motion smoothness) with a general purpose predictive model, defining the quantity of interest (e.g., flow) as the outcome of probing the model's predictions~\cite{venkatesh2023understanding}.

However, while CWM is an intriguing conceptual proposal, it has a substantial drawback: the perturbations that it relies on are hand-designed, and can be out-of-domain in real-world video. Perturbations are often not properly ``carried along'' with moving objects, resulting in suboptimal counterfactual motion extractions  (Figure~\ref{fig:perturber_details}B). As a result, the accuracy of flow extracted from CWM so far has not surpassed state-of-the-art flow estimation solutions.  

Here we present \ourmodel, a generic solution to this problem.  \ourmodels consists of two core innovations that allow us to leverage the advantages of the CWM idea while making it highly performant in real-world settings. 
First, we \emph{parameterize} a counterfactual perturbation generator with a learnable neural network (Figure~\ref{fig:perturber_details}A) that can predict perturbations specialized for the local appearance of any target points to be tracked. 
Second, we develop an approach for \emph{learning} this perturbation generator in a principled fashion without relying on any supervision from labeled data or heuristics. 
The main insight behind our approach is that the perturbation generator can be trained by applying a generalization of the asymmetric masking principle used to train the base model $\boldsymbol{\Psi}^{\texttt{RGB}}$ itself.
In particular, we connect putative flow outputs of the parameterized flow prediction function to a randomly initialized sparse flow-conditioned next-frame predictor $\boldsymbol{\Psi}^{\texttt{flow}}$ and perform joint optimization (Figure~\ref{fig:training_overview}) of both $\boldsymbol{\Psi}^{\texttt{flow}}$ and the perturbation generator.
This forces $\boldsymbol{\Psi}^{\texttt{flow}}$ to predict a future frame based on a present frame and sparse putative flow, creating an information bottleneck that generates useful gradients back on the perturbation prediction function's parameters. 

We find that \ourmodels outperforms state-of-the-art self-supervised motion estimation methods that are purposely built for this task~\cite{stone2021smurf, jiang2024doduo, shrivastava2024gmrw} when evaluated on challenging real-world benchmarks~\cite{doersch2022tap}.  We also find that \ourmodels can outperform supervised flow methods in a variety of scenarios, including examples with complex motion that are difficult to simulate with synthetic data rendering systems. The success of our approach reveals a promising direction for future work to achieve scalable counterfactual-based extraction of a variety of visual properties.

\section{Related Work}
\label{sec:relwork}

\textbf{Supervised flow estimation.} Supervised methods like RAFT~\cite{teed2020raft, wang2024sea} approach optical flow as a dense regression problem and learn from synthetic optical flow datasets~\cite{Butler:ECCV:2012, MIFDB16}. They also typically use task-specific architectures that are tailor-made for optical flow estimation, with strong inductive biases (e.g., iterative flood-filling) and task-specific regularizations to ensure learning from limited training datasets. While these methods show strong performance in many contexts, their reliance on synthetic supervision and specialized architectures limits their generalizability. It is for this reason that our self-supervised \ourmodel, which can be trained on unlimited in-the-wild videos, can outperform even supervised methods in certain key contexts.

\noindent
\textbf{Self-supervised flow with photometric loss.}
Methods for self-supervised flow learning~\cite{jonschkowski2020matters, stone2021smurf, liu2020learning}, such as SMURF~\cite{ stone2021smurf}, 
learn dense visual correspondence by optimizing photometric loss. Because of the weakness of pure photometric loss alone as supervision, these methods rely on a complex variety of heuristically chosen regularization losses (e.g., spatial smoothness of flow, among others) to achieve reasonable performance levels. Because these heuristics need to be tuned in a dataset-specific manner, these methods have failure models in complex dynamic scenes, especially with variable and large time-frame gaps. In contrast to these methods, \ourmodels does not rely on such heuristics, as the quality of the flow extraction is directly correlated with the prediction learning objective. 

\noindent
\textbf{Augmenting self-supervised flow with visual pre-training.}
A variety of methods augment photometric-based self-supervised flow using features derived from self-supervised visual pre-training~\cite{bardes2023mc,bian2022learning,xu2021rethinking,jiang2024doduo}. For example, the recent state-of-the-art Doduo method~\cite{jiang2024doduo} uses DinoV2 features as a basis on which to compute feature correspondences for downstream photometric loss.  This approach allows the extension of these methods to wider video training datasets (such as Kinetics), and thereby improves performance and generalizability.  Even when backed by strong image features, photometric loss is a comparatively weak constraint, again requiring the use of heuristic regularizers. \ourmodel, which again does not use scenario-specific heuristics, compares favorably to these methods. 

\noindent
\textbf{Point tracking.}
Point tracking across multiple frames is a related problem to flow and occlusion estimation. The majority of solutions for point tracking are supervised~\cite{harley2022particle, doersch2022tap} or semi-supervised~\cite{karaev2024cotracker3, doersch2024bootstap}, and as such are further out of scope for this work. However, several recent works propose self-supervised approaches to finding temporal correspondence typically relying on pre-trained representations~\cite{bian2022learning, jabri2020space}. These methods then extract point tracks through consistency objectives such as cycle consistency ~\cite{bian2022learning, jabri2020space, shrivastava2024gmrw} or heuristics like softmax-similarity~\cite{vondrick2018tracking} applied at the frame pair level. The current state-of-the-art in this domain, GRMW~\cite{shrivastava2024gmrw}, the main baseline comparison for our proposed \ourmodel, uses cycle consistency to build tracks based on frame pair-level predictions.

\noindent
\textbf{Real-world motion benchmarks.}
The TAP-Vid benchmark~\cite{doersch2022tap} provides a critical set of metrics for measuring the accuracy of motion-estimation systems in real-world video.  This is critical for ensuring that potential advances in motion estimation are tested against the challenges of real-world motion complexities, covering scenarios not encountered in synthetic benchmarks (e.g., non-rigid, highly articulated, deformable and breakable objects,  fluids, inelastic collisions, animate objects, and human interactions).  While originally intended for the supervised point tracking domain, recent self-supervised tracking works have begun to utilize TAP-Vid as a main benchmark for motion estimation~\cite{jiang2024doduo,shrivastava2024gmrw}. In this work, we also follow this practice. 
\section{Methods}
\subsection{Counterfactual World Modeling}
\label{sec:cwm}

\textbf{RGB-Conditioned Next Frame Predictor.} We begin by providing some background on Counterfactual World Modeling (CWM)~\cite{bear2023unifying, venkatesh2023understanding}. The first element of CWM is an RGB-conditioned next frame predictor $\boldsymbol \Psi ^\texttt{RGB}$, consisting of an encoder $\boldsymbol \Psi ^\texttt{RGB}_{\text{enc}}$ and decoder $\boldsymbol \Psi ^\texttt{RGB}_{\text{dec}}$, similar to a VideoMAE~\cite{tong2022videomae}, but trained with a 
highly asymmetric masking policy that reveals all patches of the first frame and a small fraction of patches of the second frame (see Figure~\ref{fig:background_figure}A).  Specifically, let $I_1, I_2 \in \mathbb{R}^{3\times H\times W}$ be two frame pairs in a video,
and define $\mathcal{M}_{\alpha}$ as a masking function that randomly masks some fraction, $\alpha$, of patches in an image. Given a fully visible first frame $I_1$ and a partially visible second frame $\mathcal{M}_{\alpha}(I_2)$, $\boldsymbol \Psi ^\texttt{RGB}$ is trained to predict $I_2$ by minimizing 
\begin{equation}
\begin{gathered}
\mathcal{L} = \text{MSE}(\hat{I}_2, I_2 ), \text{ where } \hat{I}_2 = \boldsymbol \Psi ^\texttt{RGB}\big(I_1, \mathcal{M}_{\alpha}(I_2)\big).
\end{gathered}
\end{equation}

\begin{figure*}[t]
    \centering
    \includegraphics[width=\linewidth]{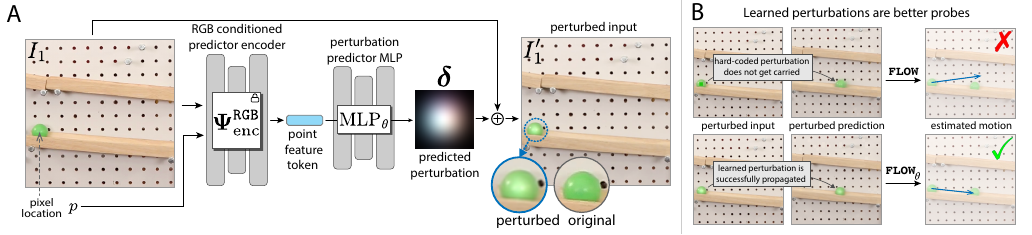}
    \vspace{-20pt}
    \caption{\textbf{Parameterizing the counterfactual intervention policy as an input-conditioned function.} (\textbf{A}) Building on a pre-trained RGB-conditioned predictor $\boldsymbol{\Psi}^\texttt{RGB}$, \ourmodels uses an image-conditioned perturbation prediction function $\delta_\theta$ containing a small MLP$_\theta$. As illustrated in \textbf{B}, $\delta_\theta$ can learn to predict image-conditioned perturbations that blend naturally with the underlying scene, potentially allowing for the perturbation to be accurately carried over to the next frame prediction. But how should the parameters of $\delta_\theta$ be learned to achieve this, without any flow supervision labels? See Figure~\ref{fig:training_overview}.
    \label{fig:perturber_details}}
    \vspace{-15pt}
\end{figure*}
Here we train CWM with $\alpha=0.1$ on the Kinetics dataset~\cite{kay2017kineticshumanactionvideo} with a frame gap of 150ms. (See the supplement for more details.)
As shown in \cite{bear2023unifying}, the asymmetric masking training policy forces $\boldsymbol \Psi ^\texttt{RGB}$ to separate scene appearance---which is wholly available in the first frame---from scene dynamics, the information of which is now concentrated in the sparse set of visible next frame patches. In other words, $\boldsymbol \Psi ^\texttt{RGB}$ is ``temporally factored''.

\noindent
\textbf{Motion Estimation.} 
Because it induces strong dependence on the appearance and position of the revealed patches from $I_1$ and $I_2$, temporal factoring allows the extraction of visual structure through applying counterfactual probes: small changes to the appearance or the position of visible patches. 
By measuring the predictor's response to these counterfactuals, we can easily extract useful information like object motion, segments, or shape from its representation~\cite{bear2023unifying}. 
As shown in Figure~\ref{fig:background_figure}B, using the \textbf{\texttt{FLOW}} procedure, a colored patch is placed on a moving object and its motion can be determined by finding its location in the predicted frame. 
To track some pixel location $p_1 = (\text{row}_1, \text{col}_1)$ from one frame to the next, input image $I_1$ is perturbed by adding a colored patch $\delta$ at pixel location $p_1$ to create the counterfactual input image $I_1' = I_1 + \delta$. 
Then, the next frames with and without the counterfactual perturbation are predicted:
\begin{equation}
\begin{gathered}
\hat{I}'_2 = \boldsymbol \Psi ^\texttt{RGB}\big( I_1 + \delta, \mathcal{M}_\alpha(I_2)\big)  = \boldsymbol \Psi ^\texttt{RGB}\big(I_1', \mathcal{M}_\alpha(I_2)\big), \\ 
\hat{I}_2 = \boldsymbol \Psi ^\texttt{RGB}\big(I_1, \mathcal{M}_\alpha(I_2)\big).
\label{second-frame-preds}
\end{gathered}
\end{equation}

Subtracting these two predicted frames and taking an $L_1$-norm across the color channels produces the difference image $\boldsymbol{\Delta} = |\hat{I}'_2 - \hat{I}_2|_1^c$.
Finally, the next-frame pixel location $\hat{p}_2$ can be computed by finding the peak in the difference image: $\hat{p}_2 = \arg\max\boldsymbol\Delta$. 

\noindent
\textbf{Occlusion.} The \textbf{\texttt{OCC}} procedure is identical to \textbf{\texttt{FLOW}} up to computation of the difference image $\boldsymbol{\Delta}$ (See Figure~\ref{fig:background_figure}). However, if a patch in the first frame gets occluded in the second frame, the response to the perturbation in the difference image $\boldsymbol{\Delta}$ will be small in magnitude and diffuse in shape. Applying a simple threshold to the maximum value of $\boldsymbol{\Delta}$ allows us to determine occlusion.

\subsection{Optimizing Counterfactual Perturbations} 
\label{subsec:optim-cfac}

\textbf{The problem with hand-designed perturbations.}
While fixed hand-designed perturbations (e.g., colored squares) can sometimes be effective in probing motion with $\boldsymbol{\Psi}^\texttt{RGB}$, they are often suboptimal--both because they are \emph{out of domain} for the base predictor and, by being image- and position-independent, can be unsuited to the local image context. Anecdotally, this results in visually obvious failure cases such as the perturbation not moving with the object or being suppressed entirely. 

Using the challenging Tap-Vid benchmark (see Section \ref{sec:experiments} for more details), we empirically quantified that the original fixed hand-designed perturbations are insufficient for self-supervised motion estimation performance (see CWM results in Table~\ref{tab: main_results}).
The main requirement for a ``good'' perturbation is thus that it be sufficiently in-distribution and image-position specific to cause meaningful context-dependent changes for probing the base predictor.
But how can perturbations be redesigned for this purpose? Our solution has two basic novel components: \emph{parameterizing} an image-conditioned counterfactual generator as a differentiable function; and formulating a general-purpose self-supervised loss objective for \emph{learning} the counterfactual generation function parameters.

\subsubsection{Parameterized Perturbations}
\label{subsubsec:interventions}
We re-formulate the motion extraction procedure from Section~\ref{sec:cwm} to make it a parameterized differentiable function and introduce the functional form of a sum of colored Gaussians as a natural perturbation class. (See Figure~\ref{fig:perturber_details}A)

Let $\textbf{\texttt{FLOW}}_\theta \colon (I_1,I_2, p_1) \mapsto \hat{\varphi}$ be a per-pixel motion estimation function with learnable parameters $\theta$ that takes an image pair $(I_1,I_2)$ and a pixel location $p_1$ in $I_1$ and outputs the predicted flow $\hat{\varphi} = \hat p_2 - p_1$. Here, $\hat p_2$ is the estimated second frame pixel location.  
The procedure $\textbf{\texttt{FLOW}}_\theta$ involves multiple components: the counterfactual perturbation function, $\delta_\theta(I_1, \mathcal{M}_{\alpha}(I_2), p_1)$, which now produces variable counterfactual perturbations as a function of the frame pair and pixel location (as opposed to a fixed perturbation $\delta$, used in the standard CWM); the pre-trained, frozen, RGB-conditioned predictor, $\boldsymbol{\Psi}^\texttt{RGB}$, as utilized within the $\textbf{\texttt{FLOW}}_\theta$ program; and a ``softargmax" module to predict a pixel location using a differentiable approximation to the argmax function.

\begin{figure*}[t]
    \centering
    \includegraphics[width=1.0\linewidth]{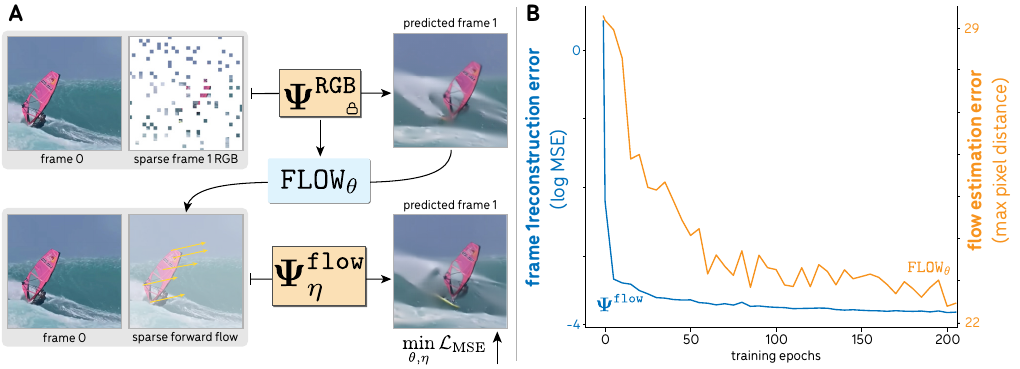}
    \vspace{-20pt}
    \caption{\textbf{A generic principle for learning optimal counterfactuals. A)} The parameterized counterfactual flow function $\textbf{\texttt{FLOW}}_\theta$ extracts motion from a frozen RGB-conditioned predictor $\boldsymbol{\Psi}^\texttt{RGB}$ through counterfactual perturbation (details in Figure~\ref{fig:perturber_details}). Its parameters $\theta$ are trained using gradients from a flow-conditioned predictor $\boldsymbol{\Psi}^\texttt{flow}_\eta$ that is jointly trained to perform next-frame prediction. The predictor $\boldsymbol{\Psi}^\texttt{flow}$ can only learn to predict future frames if it is given correct flow vectors. This explicit information bottleneck ensures useful gradients will get passed back to $\textbf{\texttt{FLOW}}_\theta$ . This setup allows us to get better extractions from a pre-trained $\boldsymbol{\Psi}^\texttt{RGB}$ predictor by training another flow-conditoned predictor $\boldsymbol{\Psi}^\texttt{flow}$ using the same principle of next-frame prediction. \textbf{(B)} As a consequence of tight coupling between the flow-conditioned predictor $\boldsymbol{\Psi}^{\texttt{flow}}$ and the learned flow estimation function $\textbf{\texttt{FLOW}}_\theta$, both motion estimation and pixel reconstruction simultaneously improve.}
    \vspace{-1.3em}
    \label{fig:training_overview}
\end{figure*}

\noindent
\textbf{Gaussian Perturbations.}  
We choose to parameterize the counterfactual perturbations as Gaussians because this function class presents a natural method of forming in-domain counterfactual input images. To compute the Gaussian parameters for a given counterfactual perturbation, we use the encoder of the RGB-conditioned predictor, $\boldsymbol \Psi ^\texttt{RGB}_{\text{enc}}$. This outputs a sequence of feature tokens from its last transformer block, which encode global and local scene content for each patch and thus form a natural basis from which Gaussian parameters can be computed using a shallow MLP. 
Given a pixel location $p_1$, we find its corresponding patch embedding token, $\boldsymbol{t}_{p_1}$, and use it as an input to an MLP that outputs a parameter vector which is in turn used to compute the Gaussian perturbation: 
\begin{equation}
    \begin{gathered}
    \boldsymbol{t}_{p_1} = \boldsymbol \Psi ^\texttt{RGB}_{\text{enc}}(I_1, \mathcal M_\alpha (I_2))_{p_1} \\
    \delta_\theta(I_1, \mathcal M_\alpha (I_2), p_1) = \text{Gaussian}\left( \text{MLP}_\theta\left(\boldsymbol{t}_{p_1}\right)\right)
    \end{gathered}
\end{equation}
Then, $\textbf{\texttt{FLOW}}_\theta$ computes the difference image, $\boldsymbol{\Delta}$, similar to the $\textbf{\texttt{FLOW}}$ program, using $\hat{I}'_2 = \boldsymbol \Psi ^\texttt{RGB}\big( I_1 + \delta_\theta, \mathcal{M}_\alpha(I_2)\big)$.
Because $\textbf{\texttt{FLOW}}_\theta$ needs to be differentiable, we use a softargmax over $\boldsymbol {\Delta}$ in place of an argmax to estimate $\hat{p}_2$.

\noindent
\textbf{Softargmax Module.} We follow the softargmax formulation proposed in~\cite{wang2020learning} to differentiably approximate the argmax function. Given a difference image, $\boldsymbol{\Delta} = |\hat{I}'_2 - \hat{I}_2|^c_1$, we first apply a temperature-scaled 2D softmax and then take the expectation according to that softmax to find the predicted second frame pixel location $\hat{p}_2 = \mathbb E_{p_2 \sim \text{softmax}(\boldsymbol{\Delta}/\tau)} [p_2]$. 
The predicted flow is then computed as 
$\hat \varphi = \hat{p}_2 - p_1$.

\subsubsection{Learning Optimized Counterfactuals}
\label{subsec: training}

Now that the perturbation generator has been parameterized, the question arises: how can its parameters be learned? What type of self-supervised objective will cause the perturbation generator function to be properly context-specific and result in accurate flow vectors? Our main insight is that this problem can be ``bootstrapped'' in a robust fashion by generalizing the sparse asymmetric mask learning paradigm to encompass a coupled and mixed-mode RGB-Flow prediction problem without using labeled data (see Figure~\ref{fig:training_overview}). Specifically, we jointly train the parameterized counterfactual motion prediction function, \textbf{\texttt{FLOW}}$_\theta$, which estimates a set of flow vectors; together with a sparse flow-conditioned predictor, $\boldsymbol \Psi ^\texttt{flow}$, which takes a single frame along with sparse flow vectors to predict the next frame. 
We constrain \textbf{\texttt{FLOW}}$_\theta$ by passing its outputs as inputs to $\boldsymbol\Psi^\texttt{flow}$ and training end-to-end using final RGB reconstruction loss on the predictions of $\boldsymbol\Psi^\texttt{flow}$.
As $\boldsymbol \Psi ^\texttt{flow}$ has no access to any RGB patches from the second frame $I_2$, it is only if the putative flows are actually correct that it be possible for $\boldsymbol\Psi ^\texttt{flow}$ to use them for minimizing next-frame reconstruction loss.  

Specifically, given an image pair $(I_1, I_2)$, we estimate the motion for a set of pixels $\mathcal{P}=\{p_1^{(1)}, p_1^{(2)}, \dots, p_1^{(n)}\}$ using $\textbf{\texttt{FLOW}}_\theta$, obtaining a set of estimated forward flow vectors $\hat {\mathcal{F}} = \{\hat \varphi^{(1)}, \hat \varphi^{(2)}, \dots, \hat \varphi^{(n)} \}$. Let $\boldsymbol \Psi ^\texttt{flow}_{\eta} \colon \big(I_1, \hat {\mathcal{F}}\big) \mapsto \hat{I}_2$ be a flow-conditioned next frame predictor with parameters $\eta$ that takes the first frame RGB input $I_1$ and predicts the next frame $\hat{I}_2$, conditioned on the flow input $\hat {\mathcal{F}}$. We jointly optimize $\theta$ and $\eta$, by minimizing $\min_{\theta,\eta} \mathcal{L}_{\text{MSE}}(\hat{I}_2, I_2)$.
Figure \ref{fig:training_overview}B shows that optimizing end-to-end reconstruction couples tightly to upstream flow accuracy, as required for effective bootstrapping.  

\noindent
\textbf{Additional Enhancements.} 
A simple random masking strategy may inadvertently reveal the ground truth RGB at the next frame location we are trying to predict for a particular point. In this event, the model will not carry over the counterfactual perturbation to the future frame, leading to an erroneous flow prediction. A simple yet effective inference-time solution is \emph{multi-mask} (MM),  in which we apply multiple random masks and average across the resulting delta images to reduce the influence of sub-optimal masks.
Following prior work~\cite{dosovitskiy2015flownet, jiang2021cotr}, we also perform an iterative multiscale refinement of flow predictions by recursively applying $\textbf{\texttt{FLOW}}_\theta$ to smaller crops centered on the predicted point location, $\hat{p}_2$ of the previous iteration. We observe that $\textbf{\texttt{FLOW}}_\theta$ is able to generate good initial flow predictions, and thus benefits from refinement (Table \ref{table:analysis}).

\begin{table*}[ht]
\renewcommand{\arraystretch}{1.4} %
\centering
{\resizebox{1.0\linewidth}{!}{
\small
\begin{tabular}{p{3mm}lccccccccccccccc}
\toprule
\multirow{2}{*}{\textbf{}} & 
\multirow{2}{*}{\textbf{Method}} & 
\multicolumn{5}{c}{\textbf{DAVIS}} & 
\multicolumn{5}{c}{\textbf{Kinetics}} & 
\multicolumn{5}{c}{\textbf{Kubric}}\\ 
\cmidrule(lr){3-7} 
\cmidrule(lr){8-12} 
\cmidrule(lr){13-17} 
& &
AJ~$\uparrow$ & AD~$\downarrow$ & $<\delta^x_\textrm{avg}$~$\uparrow$ & OA~$\uparrow$ & OF1~$\uparrow$ & 
AJ~$\uparrow$ & AD~$\downarrow$ & $<\delta^x_\textrm{avg}$~$\uparrow$ & OA~$\uparrow$ & OF1~$\uparrow$ & 
AJ~$\uparrow$ & AD~$\downarrow$ & $<\delta^x_\textrm{avg}$~$\uparrow$ & OA~$\uparrow$ & OF1~$\uparrow$ \\ 
\midrule
\multicolumn{2}{l}{\textit{TAP-Vid CFG}} \\
\midrule 

S & RAFT~\cite{teed2020raft} & 69.69 & 1.43 & 83.83 & 81.98 & 46.08 & \highlight{79.01} & \highlight{0.86} & \highlight{87.59} & 92.73 & 49.49 & 73.38 & 1.24 & 83.73 & 91.00 & 63.17 \\
& SEA-RAFT~\cite{wang2024sea} & \highlight{69.89} & 1.44 & \highlight{84.82} & 82.00 & \highlight{47.52} & 75.12 & 1.07 & 85.82 & 88.90 & 39.42 & \highlight{77.53} & \highlight{1.00} & \highlight{87.02} & \highlight{92.50} & \highlight{68.65}  \\
 
\midrule
U$^\dagger$
& Doduo~\cite{jiang2024doduo} & 25.61 & \underline{1.61} & 72.56 & 37.49 & 22.59 & 35.26 & 1.19 & 77.62 & 43.00 & 11.63 & 56.57 & 1.74 & 68.63 & 87.26 & \underline{55.01}  \\
\midrule
U
& SMURF~\cite{stone2021smurf} & \underline{65.75} & 2.40 & \underline{79.45} & 82.26 & \underline{42.65} & \textbf{78.76}& \textbf{0.97} & \textbf{87.16} & 93.13 & \underline{47.69} & \underline{69.05} & \underline{1.59} & \underline{82.38} & \textbf{90.84} & 53.49  \\
\cdashline{2-17} 
& CWM~\cite{bear2023unifying, venkatesh2023understanding} & 27.56 & 4.65 & 38.55 & \textbf{88.90} & 5.41 & 34.00 & 3.93 & 43.37 & \underline{95.17} & 5.95 & 30.72 & 4.05 & 42.33 & 88.44 & 4.27 \\
& \ourmodels (ours) & \textbf{69.53} & \textbf{1.19} & \textbf{83.15} & \underline{88.85} & \textbf{44.17} & 
\underline{75.98} & \underline{1.01} & \underline{84.31} & \textbf{96.34} & \textbf{58.61} & 
\textbf{70.70} & \textbf{1.26} & \textbf{82.78} & \underline{90.31} & \textbf{57.30}  \\
\midrule 

\multicolumn{3}{l}{\textit{TAP-Vid First --- Main Benchmark}} \\
\midrule 
S
& RAFT~\cite{teed2020raft} & 41.77 & 25.33 & 54.37 & 66.40 & 56.12 & 44.02 & 19.49 & 56.76 & 75.86 & 72.00 & 69.80 & 5.51 & 80.56 & 87.75 & 68.48 \\
& SEA-RAFT~\cite{wang2024sea} &  43.41 & 20.18 & 58.69 & 66.34 & 56.23 & 39.27 & 24.28 & 52.63 & 71.25 & 69.19 & \highlight{75.64} & 4.74 & \highlight{85.12} & \highlight{90.07} & \highlight{73.80} \\
 
\midrule
U$^\dagger$
& Doduo~\cite{jiang2024doduo} &  23.34 & \underline{13.41} & 48.50 & 47.91 & \underline{49.43} &  14.65 & \underline{16.04} & 45.84 & 45.96 & 53.94 & 51.85 & 5.67 & 64.17 & 82.65 & \underline{61.97} \\
\midrule
U
& GMRW~\cite{shrivastava2024gmrw} & \underline{36.47} & 20.26 & \underline{54.59} & 76.36 & 42.85 & 25.70 & 27.65 & 41.63 & \underline{71.33} & 31.68 & \textcolor{gray}{\underline{67.50}$^{\ddagger}$} & \textcolor{gray}{\textbf{3.16}$^{\ddagger}$}  & \textcolor{gray}{\textbf{81.74}$^{\ddagger}$} & \textcolor{gray}{\textbf{89.36}$^{\ddagger}$} &  \textcolor{gray}{35.14$^{\ddagger}$} \\
& SMURF~\cite{stone2021smurf} &  30.64 & 27.28 & 44.18 & 59.15 & 46.91 & \underline{36.99} & 28.73 & \underline{48.52} & 70.42 & \underline{64.73} & 63.47 & 6.71 & 78.78 & 87.07 & 58.60 \\
\cdashline{2-17}
& CWM \cite{bear2023unifying, venkatesh2023understanding} & 15.00 & 23.53 & 26.30 & \underline{76.63} & 18.22 & 14.84 & 30.96 & 25.00 & 70.90 & 16.79 & 26.54 & 11.81 & 39.35 & 84.14 & 13.70 \\
& \ourmodels (ours) &  \bf{47.53} & \bf{8.73} & \bf{64.83} & \bf{80.87} & \bf{60.74} & \bf{44.85} & \bf{13.44} & \bf{57.74} & \bf{84.12} & \bf{77.84} & \textbf{67.61} & \underline{4.57} & \underline{80.01} & \underline{87.95} & \textbf{67.13}  \\
\bottomrule
\end{tabular}
}}
\vspace{-0.5em}
\caption{\textbf{Quantitative results on TAP-Vid First and CFG protocols.} In the First protocol, a point is tracked from when it is first visible to the end of the video, which requires estimating motion across large frame gaps. \ourmodels outperforms both supervised and unsupervised two-frame baselines.
In the CFG protocol, point tracking is evaluated at fixed gaps of 5 frames, making it an easier setting that is more favorable to optical flow methods. ``S'' and ``U'' indicate supervised and unsupervised, respectively. Doduo is not strictly unsupervised, as explained in Section~\ref{subsec:experiments-eval-protocol}. GMRW is trained on the Kubric dataset, (marked with $\ddagger$), making it a more favorable evaluation setting for that method because of the minimal domain gap. Best performing supervised models (shaded) are considered separately.
}
\vspace{-10pt}
\label{tab: main_results}
\end{table*}

\section{Experiments}

\label{sec:experiments}
\subsection{Evaluation Protocol} \label{subsec:experiments-eval-protocol}
Our main datasets for evaluation are TAP-Vid DAVIS and TAP-Vid Kinetics~\cite{doersch2022tap}, the DAVIS~\cite{ponttuset20182017davischallengevideo} and Kinetics~\cite{kay2017kineticshumanactionvideo} datasets with human point track annotations, along with the synthetic Kubric~\cite{greff2022kubric} dataset. For TAP-Vid Kinetics and TAP-Vid Kubric, we randomly sample 30 videos (the same size as TAP-Vid DAVIS) for the evaluation. We run experiments on two distinct protocols aimed at measuring performance under various settings. For flow methods without an existing implementation of occlusion prediction, we use cycle consistency: occlusion is the event of inconsistency between forward and backward predictions greater than 6 pixels. 

Models that can accept variable resolution inputs are run with the resolution closest to native that can be fit into memory, ensuring that each is run optimally. Metrics for both procedures are always computed after rescaling predictions to $256\times256$ resolution, following~\cite{shrivastava2024gmrw}.

\noindent
\textbf{TAP-Vid First.} Following the TAP-Vid First protocol proposed in ~\cite{doersch2022tap}, for each point, we take the frame in which it is first visible and track its motion only forward in time. This is a challenging setting as it involves tracking points across variable and often large frame gaps.

\begin{figure*}[t!]
    \centering
    \includegraphics[width=\linewidth]{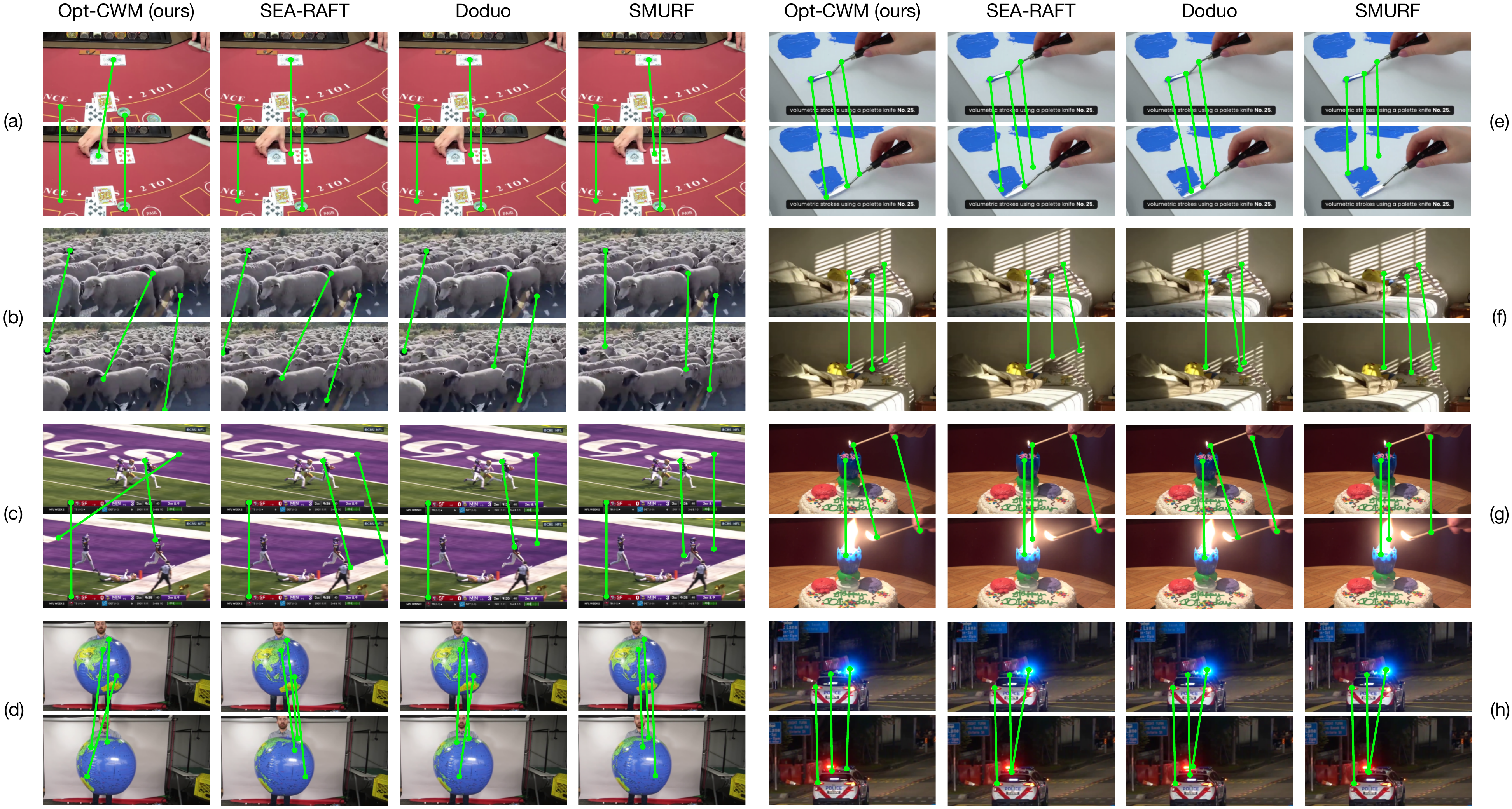}
    \vspace{-2em}
    \caption{
    \textbf{Qualitative comparison with baselines on real-world videos.} The above examples show the failure modes of previous methods that rely on visual similarity or photometric loss. We observe that the baseline models struggle against subtle but functionally important changes in largely homogeneous scenes depicting objects of similar color and texture ((a) - (e)). Further, the use of photometric loss in self-supervised methods such as SMURF can also be susceptible to differences in light intensity across frame pairs ((f) - (h)). \ourmodel, however, relies on a holistic understanding of scene transformations and object dynamics and is able to find correspondence without arbitrary heuristics. 
    }
    \label{fig:baseline_comparisons}
    \vspace{-18pt}
\end{figure*}

\noindent
\textbf{TAP-Vid Constant Frame Gap (CFG).} For fair comparison with optical flow models, we also propose an additional protocol with fixed frame gaps that is more advantageous for these baselines (see supplementary for the effect of frame gap on flow baselines). In particular, a fixed 5-frame gap is used: metrics are computed on all frame pairs that are 5 frames apart (and the point is visible in the first).

\noindent
\textbf{Metrics.} We use the official metrics from the TAP-Vid evaluation protocol~\citep{doersch2022tap}: 1) \textit{Average Jaccard} (AJ), a precision metric measuring a combination of point tracking and occlusion prediction; 2) \textit{Average Distance} (AD) between the estimated pixel and ground truth locations; 3) $< \delta^x_{avg}$, which measures the average percentage of points predicted correctly within a variety of pixel distance thresholds; and 4) \textit{Occlusion Accuracy} (OA), the fraction of points correctly predicted as either occluded or visible. Additionally, to account for the relative lack of occlusion events in the dataset, we also evaluate 5) \textit{Occlusion F1} (OF1), which computes the F1 score of the occlusion predictions. 

\subsection{Baselines}
Our evaluation protocol requires tracking points in videos through occlusion by finding temporal correspondence: given a frame pair, determine where the point went or whether it was occluded. Therefore, the appropriate baselines are supervised and self-supervised optical flow methods, and self-supervised temporal correspondence methods. We run the following baselines:

\textbf{CWM}~\cite{bear2023unifying, venkatesh2023understanding} represents motion estimated through counterfactual extractions with a fixed perturbation and without additional enhancements. This comparison illustrates the significant performance gains over prior CWM works. We present a detailed ablation analysis in Section~\ref{sec:ablation}. 

\textbf{GMRW}~\cite{shrivastava2024gmrw} is a recent self-supervised approach to video correspondence. The transformer-based architecture is trained on cycle consistency using contrastive random walks. GMRW is designed for temporal correspondence-based long-range tracking and is the SOTA baseline for comparison on TAP-Vid First. 

\textbf{SMURF}~\cite{stone2021smurf} is an unsupervised method specifically designed for optical flow estimation. This work tailors the RAFT~\cite{teed2020raft} architecture so it can be trained using optical flow-specific heuristic losses like photometric loss and smoothness regularization. SMURF specializes in estimating motion in consecutive frames, with models trained on KITTI, Sintel, and FlyingChairs.

\textbf{Doduo}~\citep{jiang2024doduo} is a method for finding dense correspondence across images trained on unlabeled, in-the-wild videos from Youtube-VOS~\cite{xu2018youtube}. It uses photometric losses and leverages the DINO~\cite{caron2021emerging} encoder for incorporating sparse correspondence priors. Doduo is not strictly unsupervised, as it uses off-the-shelf Mask2Former segments~\cite{cheng2022masked}. 

\textbf{SEA-RAFT}~\cite{wang2024sea} is a supervised flow method that builds upon the original RAFT~\cite{teed2020raft} by adding additional pretraining on TartanAir~\cite{wang2020tartanair}, a novel mixture of Laplace loss, and improved initialization of the flow estimation.

Evaluation results on TAP-Vid First and CFG are presented in Table~\ref{tab: main_results}. Our best-performing model accepts 512 resolution inputs and is evaluated with 10 multi-masks and 4 multiscale iterations (see Section \ref{subsec: training}). On TAP-Vid First, \ourmodels outperforms all baselines on all metrics.
In particular, \ourmodels greatly improves upon AD, demonstrating robustness even in difficult (though more realistic) cases with long frame gaps or high motion. 
On the CFG protocol, which is favorable to flow methods, \ourmodels either outperforms or is competitive with the best unsupervised models, especially on DAVIS. Additionally, in Figure \ref{fig:baseline_comparisons}, we show qualitative comparisons between methods. As expected, the baselines struggle with videos violating the heuristic assumptions for which they were optimized, e.g., photometric similarity. Our experiments on the synthetic Kubric dataset~
\cite{greff2022kubric}, which more similar to synthetic optical flow training datasets and therefore more favorable to methods trained on such data, demonstrate that \ourmodels has the best performance in this out of domain scenario among models that were not trained on Kubric data. Further \ourmodels qualitative examples can be found in the supplement.

\begin{figure*}[t!]
    \centering
    \includegraphics[width=\linewidth]{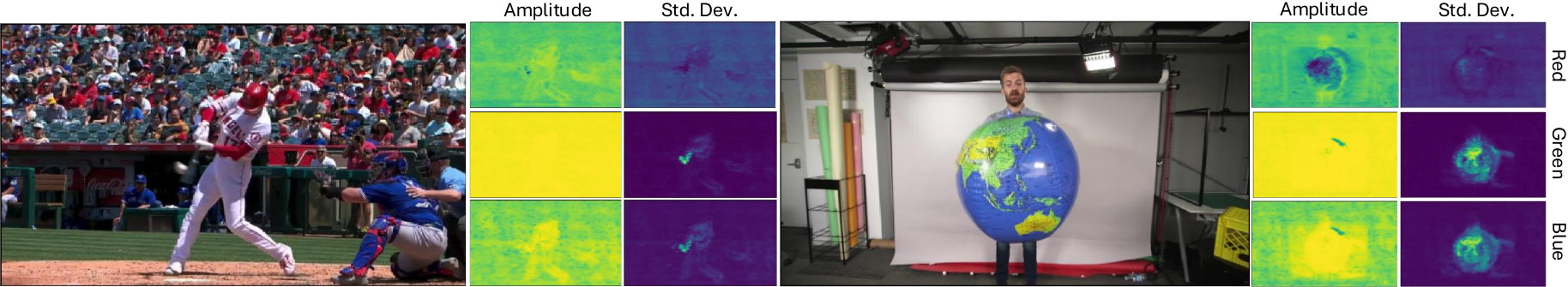}
    \vspace{-20pt}
    \caption{\textbf{Perturbation maps emergently reflect scene properties.}  For two example frame pairs, we show the amplitudes and standard deviations, at each spatial position and for each color channel, of the optimal Gaussian perturbations predicted by MLP$_\theta$. These ``perturbation maps'' emergently reflect scene properties, with perturbation parameters varying in size and magnitude depending on where they are located in the image, corresponding to the presence of foreground objects and their parts.}
    \label{fig:pert_map}
    \vspace{-10pt}
\end{figure*} 

\subsection{Ablation and Analysis}
\label{sec:ablation}

\begin{table}[t]
\centering
\scalebox{0.7}{
\renewcommand{\arraystretch}{1.1}
\begin{tabular}{cccc|ccccc}
\Xhline{0.8pt}
\Gape[5pt][0pt]
Type & MM & MS & Res. & AJ~$\uparrow$ & AD~$\downarrow$ & $<\delta^x_\textrm{avg}$~$\uparrow$ & OA~$\uparrow$ & OF1~$\uparrow$ \\ [1pt]
\hline
\Gape[5pt][0pt]
learned & 10 & 4 & 512 & \bf{47.53} & \bf{8.73} & \bf{64.83} & \underline{80.87} & \bf{60.74} \\
learned & 1 & 4 & 512 & \underline{42.85} & 9.82 & \underline{59.72} & 78.55 & \underline{60.20} \\
learned & 10 & 0 & 512 & 32.71 & 11.98 & 49.20 & 79.28 & 41.45\\
learned & 3 & 2 & 512 & 40.51 & \underline{9.72} & 58.57 & 80.34 & 50.06 \\ 
red square & 3 & 2 & 512 & 21.37 & 18.25 & 36.31 & 75.38 & 27.21 \\ 
green square & 3 & 2 & 512 & 30.44 & 12.72 & 47.37 & 76.89 & 19.10 \\
learned & 3 & 2 & 256 & 37.00 & 11.62 & 52.82 & \bf{81.10} & 57.84  \\ 
learned & 1 & 0 & 256 & 21.85 & 20.55 & 34.34 & 78.03 & 53.10 \\
red square & 1 & 0 & 256 & 15.00 & 23.53 & 26.30 & 76.63 & 18.22\\
green square & 1 & 0 & 256 & 19.91 & 19.61 & 32.73 & 78.31 & 36.53\\
[1pt]
\hline 
\end{tabular}
}
\caption{\textbf{Ablation analysis TAP-Vid DAVIS First protocol.} We evaluate multi-mask (MM) and multiscale (MS), in addition to comparing our optimized perturbations (``learned") with the fixed ones (``red square"/``green square")~\citep{bear2023unifying, venkatesh2023understanding}. MM and MS columns indicate the number of masking or zooming iterations. We observe a clear improvement on all metrics, highlighting the need for bespoke, in-distribution counterfactual perturbations, multi-mask inference and multi-scale refinement.}
\label{table:analysis}
\vspace{-15pt}
\end{table}



\textbf{Optimizing Counterfactuals.}
We compare Opt-CWM with a spectrum of types of hard-coded perturbations, representing various forms of unoptimized CWM baseline, and find that learned interventions perform substantially better (see Table \ref{table:analysis}). This demonstrates not only that the CWM framework is highly effective at unsupervised motion estimation, but also that optimizing counterfactual perturbations is critical for good performance. The highly image-dependent nature of the optimal predicted perturbations is illustrated in Figure~\ref{fig:pert_map}, which shows variations in Gaussian parameters of the perturbation as a function of where the perturbation is to be placed in the image. As shown in the supplement, this scene-dependent perturbation is an emergent result of our training procedure that directly contributes to improvements in flow predictions.

\noindent\textbf{Resolution and additional refinements.}
In Table~\ref{table:analysis}, we also observe that models with a larger input resolution outperform those with a smaller one, and that our multi-mask inference (MM) and multiscale refinement (MS) procedures significantly improve performance.

\subsection{Distillation into a RAFT architecture}
RAFT, SEA-RAFT, and SMURF use a highly efficient but special-purpose flow architecture, rather than large general purpose ViTs~\cite{dosovitskiy2020image}. To isolate the effect of this specific architecture design, we train a RAFT-type architecture with pseudo-labels generated by \ourmodel. Specifically, we take a frame pair for each clip pseudo-label the motion for 1\% of the pixels, and train a SEA-RAFT architecture on this pseudo-labeled dataset. We present our findings in Table~\ref{tab:consolidated}, and find that this distilled model outperforms the equivalent self-supervised baseline SMURF and is competitive with the supervised techniques. This outcome pinpoints that the core reason for \ourmodel's improved performance is our contribution of the novel optimized counterfactual extraction scheme, and the flexible ability to train on unrestricted data that this approach enables, rather than the ViT architecture as such.  

\begin{table}[ht]
\centering
\resizebox{\linewidth}{!}{%
\renewcommand{\arraystretch}{1.1}
\small
\begin{tabular}{p{3mm}lccccc}
\toprule
\multirow{2}{*}{\textbf{}} & 
\multirow{2}{*}{\textbf{Method}} & 
\multicolumn{5}{c}{\textbf{DAVIS}} \\ 
\cmidrule(lr){3-7} 
& &
AJ~$\uparrow$ & AD~$\downarrow$ & $<\delta^x_\textrm{avg}$~$\uparrow$ & OA~$\uparrow$ & OF1~$\uparrow$ \\ 
\midrule
\multicolumn{2}{l}{\textit{TAP-Vid CFG}} \\
\midrule 
S & RAFT~\cite{teed2020raft} & 69.69 & 1.43 & 83.83 & 81.98 & 46.08 \\
& SEA-RAFT~\cite{wang2024sea} & 69.89 & 1.44 & \highlight{84.82} & 82.00 & 47.52 \\
\midrule
U & SMURF~\cite{stone2021smurf} & 65.75 & 2.40 & 79.45 & 82.26 & 42.65 \\
\noalign{\vskip 1mm}
\cdashline{2-7} 
\noalign{\vskip 1mm}
& \ourmodel & \underline{69.53} & \textbf{1.19} & \textbf{83.15} & \textbf{88.85} & \underline{44.17} \\
& \ourmodels Distilled & 
\textbf{70.51} & \underline{1.30} & \underline{82.11} & \underline{88.05} & \textbf{55.04} \\
\midrule 
\multicolumn{7}{l}{\textit{TAP-Vid First --- Main Benchmark}} \\
\midrule 
S & RAFT~\cite{teed2020raft} & 41.77 & 25.33 & 54.37 & 66.40 & 56.12 \\
& SEA-RAFT~\cite{wang2024sea} & 43.41 & 20.18 & 58.69 & 66.34 & 56.23 \\
\midrule
U & SMURF~\cite{stone2021smurf} & 30.64 & 27.28 & 44.18 & 59.15 & 46.91 \\
\noalign{\vskip 1mm}
\cdashline{2-7}
\noalign{\vskip 1mm}
& \ourmodel & \bf{47.53} & \bf{8.73} & \bf{64.83} & \bf{80.87} & \bf{60.74} \\
& \ourmodels Distilled & \underline{44.05} & \underline{17.49} & \underline{57.93} & \underline{69.75} & \underline{60.72} \\
\bottomrule
\end{tabular}%
}
\vspace{-0.5em}
\caption{\textbf{\ourmodels Distillation Results.} To obtain fast test-time inference with a small model, we distill \ourmodels into the smaller and more efficient SEA-RAFT architecture by sparsely pseudo-labeling Kinetics with \ourmodel. This approach outpeforms the self-supervised SMURF and is competitive with the supervised RAFT models, while requiring no labeled training data.
}
\vspace{-15pt}
\label{tab:consolidated}
\end{table}

\section{Conclusion \& Future Work}
We have demonstrated the effectiveness of \ourmodels in understanding motion concepts, achieving state-of-the-art performance on real-world benchmarks. Our paper takes an essential first step in demonstrating the potential of optimized counterfactuals for probing pre-trained video predictors. \ourmodels paves the way for the next generation of scalable self-supervised point trackers, with potential extensions by training multi-frame predictors, scaling training data, or using different base predictor architectures like autoregressive generative video models. Equally importantly, our twin techniques of parameterizing the input-conditioned counterfactual generator and bootstrapping the learning of the generator parameters with end-to-end sparse prediction loss are generic and not flow-specific---and may thus be extensible to optimizing highly performant CWM-style extraction of a wide variety of visual quantities, including object segments, depth maps, and 3D shape~\cite{bear2023unifying,venkatesh2023understanding}.

\section{Acknowledgements}
This work was supported by the following awards: To D.L.K.Y.: Simons Foundation grant 543061, National Science Foundation CAREER grant 1844724, National Science Foundation Grant NCS-FR 2123963, Office of Naval Research grant S5122, ONR MURI 00010802, ONR MURI S5847, and ONR MURI 1141386 - 493027. We also thank the Stanford HAI, Stanford Data Sciences and the Marlowe team, and the Google TPU Research Cloud team for computing support.

{
    \small
    \bibliographystyle{ieeenat_fullname}
    \bibliography{main}
}

\clearpage
\appendix

\section{Implementation Details}

\subsection{Architecture Details}

\subsubsection{\texorpdfstring{$\boldsymbol{\Psi}^{\texttt{RGB}}$}{Psi RGB}}

\label{figmention:figure_1}
The input video is first divided into non-overlapping spatiotemporal patches of size $8\times8$, with a subset of patches masked. Unlike MAE, we train with both revealed input patches and mask tokens provided to the encoder. We train with the ViT-B architecture~\cite{he2022masked} with each transformer block consisting of a multi-head self-attention block and an MLP block, both using LayerNorm (LN). The CWM decoder has the same architecture as the encoder. Each spatiotemporal patch has a learnable positional embedding which is added to both the encoder and decoder inputs. CWM does not use relative position or layer scaling~\cite{bao2021beit, he2022masked}. Please refer to~\cite{venkatesh2023understanding, bear2023unifying} for more details on the architecture

\noindent\textbf{Default settings} We show the default pre-training settings in Table \ref{table:pre_training_setting}. CWM does not use color jittering, drop path, or gradient clip. Following ViT’s official code, Xavier uniform is used to initialize all transformer blocks. The learnable masked token is initialized as a zero tensor. Following MAE, we use the linear lr scaling rule: $lr = base\_lr\times batch\_size\,/\,256$ \cite{he2022masked}.

\begin{table}[ht]
\centering

\caption{\textbf{Default pre-training setting of CWM}}
\begin{tabular}{l|l}
\toprule
\textbf{config}          &  \,value\\ \midrule
optimizer                & \,AdamW \cite{loshchilov2017decoupled}                         \\
base learning rate       & \,1.5e-4                        \\
weight decay             & \,0.05                          \\
optimizer momentum       & \,$\beta_1, \beta_2=0.9, 0.95$ \cite{chen2020generative}  \\
accumulative batch size         & \,4096                          \\
learning rate schedule   & \,cosine decay  \cite{loshchilov2016sgdr}                \\
warmup epochs \cite{goyal2017accurate}           & \,40                            \\
total epochs             & \,800                          \\
flip augmentation        & \,no                            \\
augmentation             & \,MultiScaleCrop  \cite{wang2018temporal}       \\ \bottomrule
\end{tabular}
\label{table:pre_training_setting}
\vspace{-0.5cm}
\end{table}


\subsubsection{\texorpdfstring{$\boldsymbol{\Psi}^{\texttt{flow}}$}{Psi flow}}
The architecture of the flow-conditioned predictor, $\boldsymbol{\Psi}^{\texttt{flow}}$, is a vision transformer with 16 layers and 
132M parameters. Input images are resized to 224x224, and the patch size is 8. Sinusoidal positional encodings are used.
For the encoder, the embedding dimension is 768, and 12 attention heads are used. For the decoder, the embedding dimension is 384, and 6 attention heads are used. 

This model has two parallel ``streams", the first of which takes RGB input and the second of which takes sparse flow, concatenated with RGB (which is masked to have the same sparsity as the flow), as input. All RGB inputs are from the first frame only; this requires the model to depend solely on flow to modify the RGB and predict the next frame.

The transformer architecture then applies self-attention to each stream and cross-attention between streams. The encoder has 12 layers, split into three groups of 4. In each group, there is one layer with self attention on each stream and cross attention from each stream to the other, followed by three layers with only self-attention on the first stream. The decoder has 4 layers; the first applies self-attention to each stream and cross-attention from each stream to the other; the second applies self-attention to the first stream and cross-attention from the second stream to the first; and the final two only apply self-attention to the first stream.



\subsection{Training Details}

\subsubsection{\texorpdfstring{$\boldsymbol{\Psi}^{\texttt{RGB}}$}{Psi RGB}}
We train CWM at 256 resolution for 800 epochs and finetune at 512 resolution for 100 epochs by interpolating the positional embeddings. It takes approximately 4 days to train 800 epochs on a TPU v5-128 pod. We pre-train CWM on the Kinetics-400 dataset \cite{kay2017kineticshumanactionvideo}, without requiring any specialized temporal downsampling. 

Applying the temporal masking strategy, i.e. fully revealing the first frame and partially revealing the second frame, during the pre-training of \psirgb \ contributes significantly to downstream flow prediction quality, compared to the baseline Video-MAE-style masking policies (Table~\ref{table:train-mr-analysis}). The asymmetry forces \psirgb \ to separate scene appearance from dynamics, as discussed in Section 3.1 in the main text.
\begin{table}[t]
\centering
\scalebox{0.7}{
\renewcommand{\arraystretch}{1.1}
\begin{tabular}{lll|ccccc}
\Xhline{0.8pt}
\Gape[5pt][0pt]
Mask Type & Train \% & Test \% & AJ~$\uparrow$ & AD~$\downarrow$ & $<\delta^x_\textrm{avg}$~$\uparrow$ & OA~$\uparrow$ & OF1~$\uparrow$ \\ 
\hline
\Gape[5pt][0pt]
tube & 55-55 & 0-90 & 23.94 & 15.61 & 36.90 & 72.19 & 52.36 \\
tube & 75-75 & 0-90 & 22.55 & 15.86 & 39.63 & 58.20 & 52.27 \\
tube & 90-90 & 0-90 & 15.23 & 18.57 & 32.12 & 51.98 & 49.20 \\
random & 75-75 & 0-90 & 29.09 & 14.64 & 42.57 & 73.51 & 57.06 \\
random & 75-75 & 0-75 & 34.06 & 12.79 & 47.54 & 76.07 & 60.81 \\
\hline 
random & 0-90 & 0-90 & 37.00 & 11.62 & 52.82 & 81.10 & 57.80 \\
\hline 
\end{tabular}
}
\caption{\textbf{Ablation analysis of training-time masking policy on TAP-Vid DAVIS First.} We train $\boldsymbol{\Psi}^{\texttt{RGB}}$ with different non-temporally factored masking policies more similar to Video-MAE~\cite{tong2022videomae, wang2023videomae}. The notation of 55-55 indicates 55\% of patches are masked in the first frame and 55\% are masked in the second frame. Tube masking selects patches at the same spatial location over time, whereas random independently samples patches in each frame. MAE-style masking during training is strictly worse than the temporally-factored masking policy we use as the standard in this paper (shown for reference in the bottom row). All experiments here use 256x256 resolution, MM-3 and MS-2.}
\label{table:train-mr-analysis}
\vspace{-5pt}
\end{table}

\subsubsection{\texorpdfstring{$\boldsymbol{\Psi}^{\texttt{flow}}$ and $\texttt{FLOW}_\theta$}{Psi flow and Flow theta}}
We train $\boldsymbol{\Psi}^{\texttt{flow}}$ and $\texttt{\textbf{FLOW}}_\theta$ jointly using an AdamW optimizer with weight decay of $0.05$, betas of $(0.9,0.95)$, and a learning rate schedule with max learning rate $1.875 \times 10^{-5}$, 40 warmup epochs (10\% of total training epochs), and cosine decay. We used a batch size of 32.


\subsection{Inference Techniques}

\subsubsection{Multi-Mask}
In the process of computing flows in $\texttt{\textbf{FLOW}}_\theta$, at inference time, we take an argmax over the difference between the predicted next frame with and without the counterfactual perturbation. This difference image, $\Delta$, depends on the choice of the random mask as this mask is used by $\boldsymbol \Psi^\texttt{RGB}$ for the next-frame reconstruction. As discussed in the main text, if a random mask reveals patches too close to where the perturbation should be reconstructed, the predictor $\boldsymbol{\Psi}^\texttt{RGB}$ may not reconstruct the perturbation properly, and the difference image will be noisy and diffuse, preventing the model from accurately predicting the next-frame location. Additionally, the reconstructed pixels will not necessarily be the same across different random samplings of visible patches, which may add random noise to the difference image. Both of these issues are ameliorated by our multi-mask technique, in which we compute difference images for a variety of sampled random masks (we found 10 to be a good number of masks for multi-masking), average over the difference images, and then take the argmax of this averaged $\Delta_\text{avg}$ to compute next-frame location for determining flow.

\subsubsection{Multiscale}
\label{subsec:multiscale}

\begin{figure}[t]
    \centering
    \includegraphics[width=\linewidth]{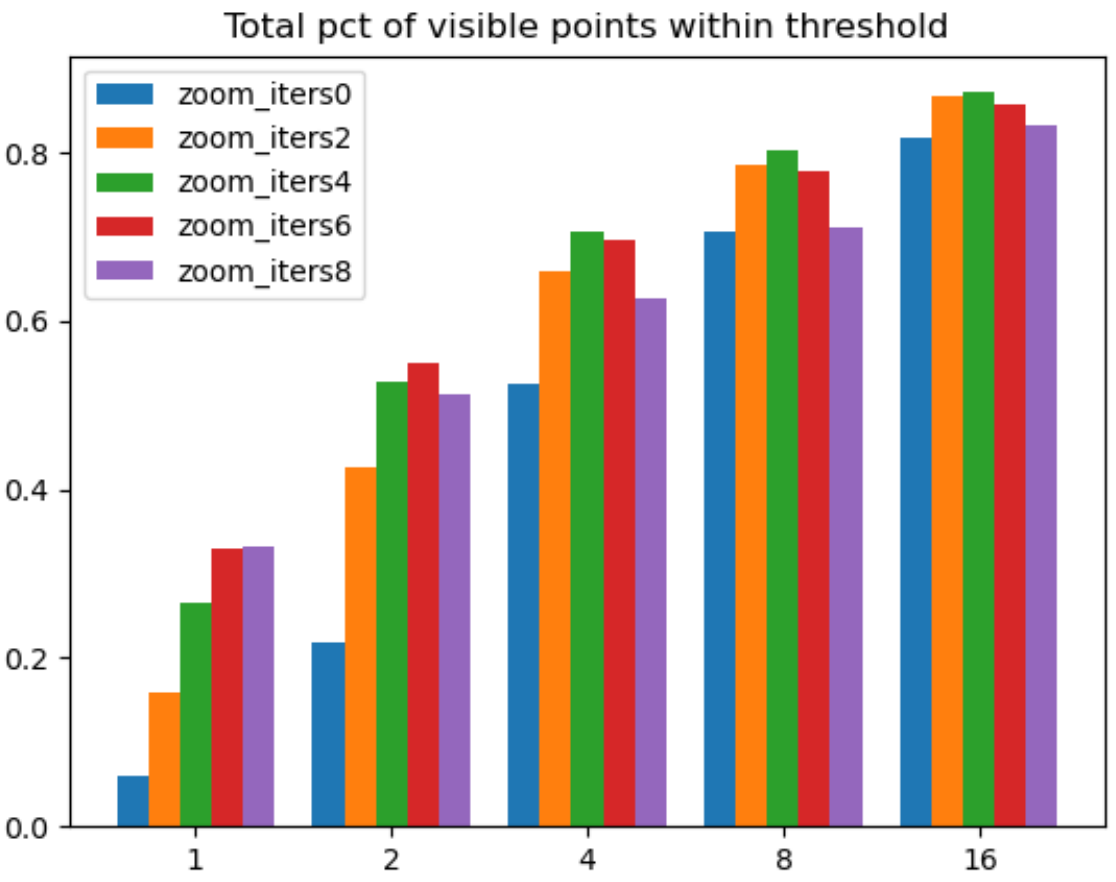}
    \caption{\textbf{$<\boldsymbol{\delta}_\text{avg}$ broken down across thresholds ($x$-axis).} Fraction of points with error less than a fixed threshold, as a function of number of multiscale (MS) iterations, for pixel thresholds 1, 2, 4, 8, and 16. We find that 4 zoom iterations tends to perform the best, especially for robustness on difficult examples (evidenced by better performance on higher thresholds).}
    \label{fig:zoom_iters}
\end{figure} 

Multiscale refinement of the original flow prediction improves \ourmodel's performance, as observed in Figure~\ref{fig:zoom_iters}. Given an input frame pair and an initial flow prediction, we perform iterative multiscaling through the following procedure. At each ``zoom iteration'', we take a $0.75H \times 0.75W$ crop of the input frames with original height $H$ and width $W$. We center the second frame crop on the location predicted by the previous iteration.

The transformer-based architecture of the next frame predictor $\boldsymbol{\Psi}^{\texttt{RGB}}$ imposes a limit to the input resolution, which may occasionally prevent small objects or minute features of the input frame from being accurately reconstructed in great detail. Multiscale refinement of the initial flow prediction can be greatly beneficial under these circumstances. However, Figure~\ref{fig:zoom_iters} suggests that the improvement is not monotonic; indeed, excessive cropping may lead to the loss of global context that is necessary to accurately reconstruct the scene. \ourmodels is run on 4 zoom iterations, which we have empirically found to be optimal.

\subsubsection{Occlusion Estimation}
The difference image $\Delta$ can also be used to predict whether a visible point becomes occluded in the next frame. Conceptually, as described in Section 3 in the main text, when a point becomes occluded, the counterfactual perturbation placed on the object should not be reconstructed in the second frame. Thus, while we take argmax $\Delta$ to compute flow, we can instead use $\max \Delta$ as a signal for occlusion. In particular, we compare $\max \Delta$ to some threshold $t_\text{occ}$ to predict occlusion (i.e., we consider the model to have predicted that a point becomes occluded if and only if $\max \Delta < t_\text{occ}$).

In the multi-masking setting with 10 masking iterations, we have 10 difference images: $\Delta_1$, $\Delta_2$, ..., $\Delta_{10}$. Instead of thresholding the average, $\Delta_\text{avg}$, we can get an improved signal by considering $\max \Delta_i$ for each $i = 1,...,10$. In this setting, we found that checking $\frac 1{10} \sum_{i=1}^{10} \max \Delta_i < 0.05$ provided a good signal for predicting occlusion, and this prediction criterion is what was evaluated in the OA and OF1 metrics of Table 1 in the main paper.

\section{Additional Quantitative Results}
\subsection{Masking Ratio at Inference}
\begin{table}[t]
\centering
\scalebox{0.7}{
\renewcommand{\arraystretch}{1.1}
\begin{tabular}{lcccccc}
\toprule
Masking Ratio & AJ~$\uparrow$ & AD~$\downarrow$ & $<\delta^x_\textrm{avg}$~$\uparrow$ & OA~$\uparrow$ & OF1~$\uparrow$ \\ [1pt]
\midrule
50\% & 42.78 & 10.52 & 58.78 & 79.18 & \underline{60.68} \\
60\% & \bf{43.28} & 10.12 & 59.56 & 80.33 & \bf{60.80} \\
70\% & \underline{43.25} & 9.72 & \bf{59.95} & \underline{81.24} & 59.68 \\
80\% & 42.68 & \bf{9.44} & \underline{59.76} & \bf{81.64} & 57.53 \\
85\% & 41.99 & \underline{9.57} & 59.58 & 80.92 & 54.06 \\
90\% (Ref.) & 40.51 & 9.72 & 58.57 & 80.34 & 50.06 \\
95\% & 37.68 & 10.57 & 55.87 & 79.63 & 45.00 \\
98\% & 32.85 & 13.15 & 50.48 & 78.19 & 41.42 \\

\bottomrule
\end{tabular}
}

\caption{\textbf{Ablation analysis of test-time masking policy on TAP-Vid DAVIS First.} We evaluate a 512 resolution $\boldsymbol{\Psi}^{\texttt{RGB}}$ across various masking ratios for the second frame using the MM-3 and MS-2 setting. The standard masking ratio for all results in this work is included as 90\% (Ref.) in this table.}
\label{table:inference-mr-analysis}
\vspace{-10pt}
\end{table}

The amount of next-frame information to provide \psirgb \ when making counterfactual predictions is an important hyperparameter for downstream performance. Intuitively, decreasing the masking ratio (i.e. revealing more next-frame patches) will improve reconstruction quality. This can improve flow prediction by focusing the source of noise in the delta image on the carried over perturbation (and not on spurious noise induced by poor reconstruction quality). On the other hand, revealing too many patches may uncover the ground truth appearance of the perturbed patch in the next frame, in which case the perturbation will not be reconstructed at all. We investigate this tradeoff in Table~\ref{table:inference-mr-analysis}. 
\subsection{Performance as a Function of Frame Gap}
\begin{figure*}[t]
    \centering
    \includegraphics[width=\linewidth]{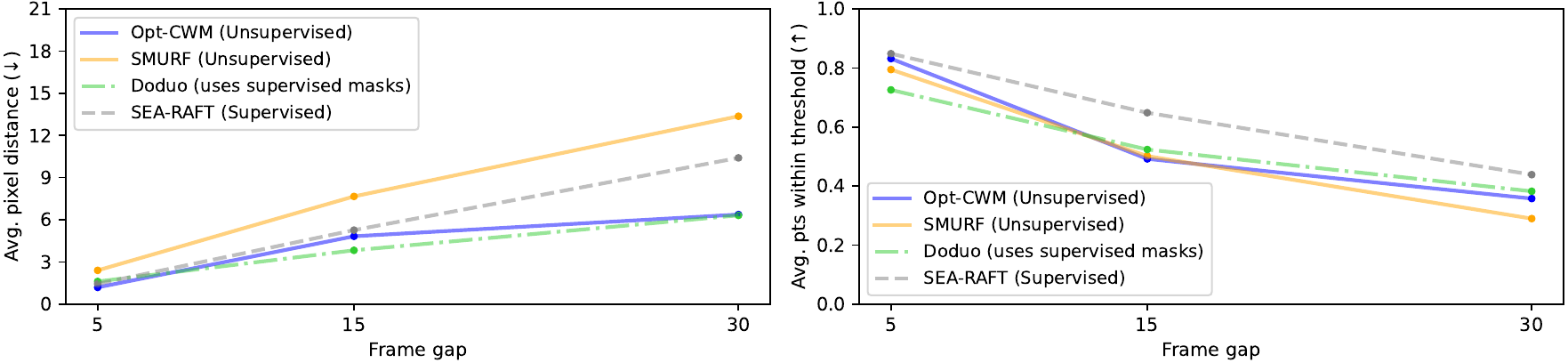}
    \caption{\textbf{Model comparison as a function of frame gap.} Higher frame gaps present harder flow estimation problems due to including more motion, as reflected by improved performance across models in lower frame gap settings. \ourmodels and Doduo perform better as frame gap increases, in contrast to optical flow methods SEA-RAFT and SMURF which decay in performance as motion magnitude increases, especially on the AD metric.}
    \label{fig:frame_gap}
\end{figure*}

\begin{figure}[t]
    \centering
    \includegraphics[width=\linewidth]{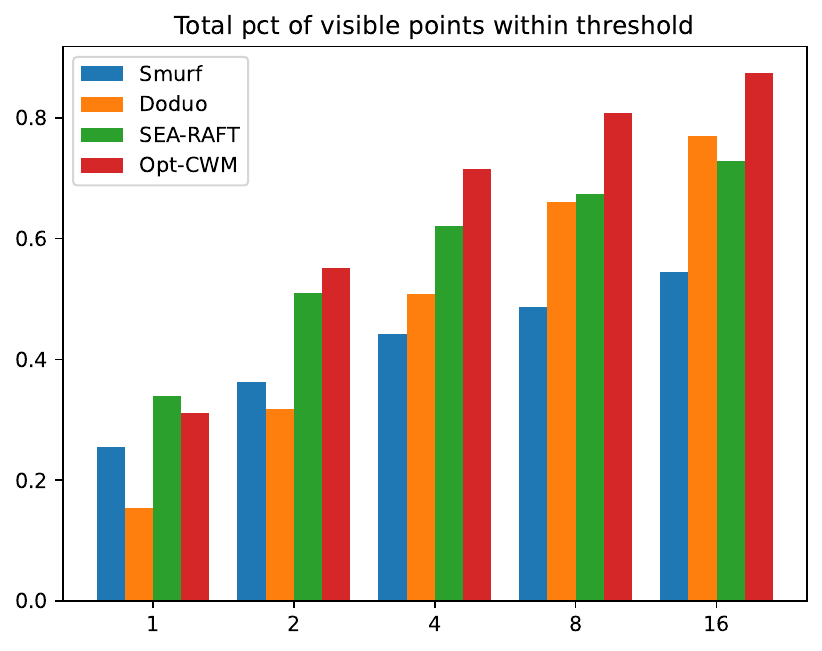}
    \caption{\textbf{TAP-Vid First: comparing baseline models on $<\boldsymbol{\delta}_\text{avg}$ broken down across thresholds ($x$-axis).} Fraction of points with error less than a fixed threshold, as a function of baseline model. Compared to baseline models, \ourmodels maintains high performance on all thresholds even when making predictions across large frame gaps, as is necessary for TAP-Vid First.}
    \label{fig:precision}
\end{figure}

\begin{figure}[t]
    \centering
    \includegraphics[width=\linewidth]{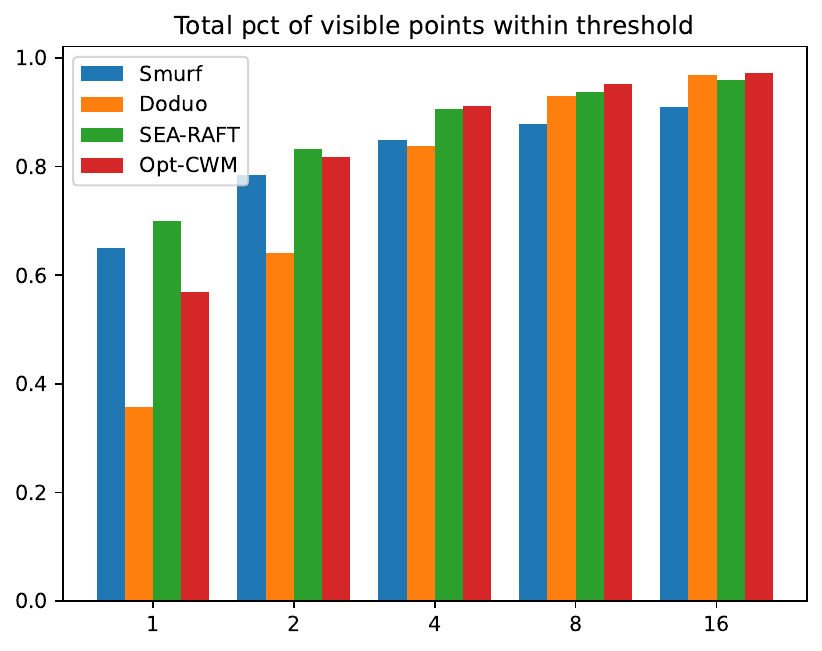}
    \caption{\textbf{TAP-Vid CFG: comparing baseline models on $<\boldsymbol{\delta}_\text{avg}$ broken down across thresholds ($x$-axis).} Fraction of points with error less than a fixed threshold, as a function of baseline model. For fair comparison, we also evaluate on a constant frame gap setting that is more favorable to optical flow baselines. While baseline methods show strong performance for very low thresholds ($< 2$ pixels), we see that in general \ourmodels outperforms self-supervised methods and is comparable with SEA-RAFT in predicting more points within a reasonable boundary.}
    \label{fig:precision_cfg}
\end{figure} 

Figure \ref{fig:frame_gap} compares flow estimation performance as a function of frame gap for \ourmodels and three baselines. \ourmodels and Doduo maintain performance as frame gap increases more than SEA-RAFT or SMURF. Doduo is trained on larger frame gaps (1-3 seconds) than SEA-RAFT, SMURF, or \ourmodels ($< 500$ms), which accounts for its strong performance as frame gap increases. Despite this difference, \ourmodels is competitive with Doduo on larger frame gaps. TAP-Vid First results are averaged over frame gaps ranging from 1 to the length of the clip (which can extend up to 50-100 frames depending on the video), and \ourmodels significantly outperforms Doduo on all metrics. Larger frame gaps often entail greater magnitudes of object and camera motion, and therefore \ourmodel's high performance as evidenced in Figure~\ref{fig:frame_gap} suggests a robustness to challenging scene dynamics. 


\subsection{Precision Analysis}

Figure~\ref{fig:precision} attempts to explain the high performance of \ourmodels on TAP-Vid First through a similar analysis done in Section~\ref{subsec:multiscale}. Our best-performing model (with optimal inference-time configurations) is able to predict the next frame location within 16 pixels of the ground truth for over 85\% of the total number of visible points. Unlike baseline models, \ourmodels is able to predict most points within a reasonable boundary. Further, \ourmodels predictions are precise; it predicts the majority of the query points within 2 pixels of the ground truth. While SEA-RAFT, which is supervised, is also precise for lower thresholds, the magnitude of the error for wrong predictions is evidently higher, as its performance quickly plateaus for higher thresholds.

As discussed in Section 4 in the main paper, we further evaluate on a custom constant-frame gap protocol (CFG) for fairer comparison with optical flow baselines. As shown here in Figure~\ref{fig:precision_cfg}, all models improve significantly under this less challenging setup. In particular, optical flow baselines exhibit strong sub pixel precision. However, we see that in general, compared to self-supervised baselines, \ourmodels make reasonable predictions of a point's next frame location more often, at a rate comparable to the fully supervised SEA-RAFT.


\subsection{Perturbation Across Epochs}

\begin{figure}[ht]
    \centering
    \includegraphics[width=\linewidth]{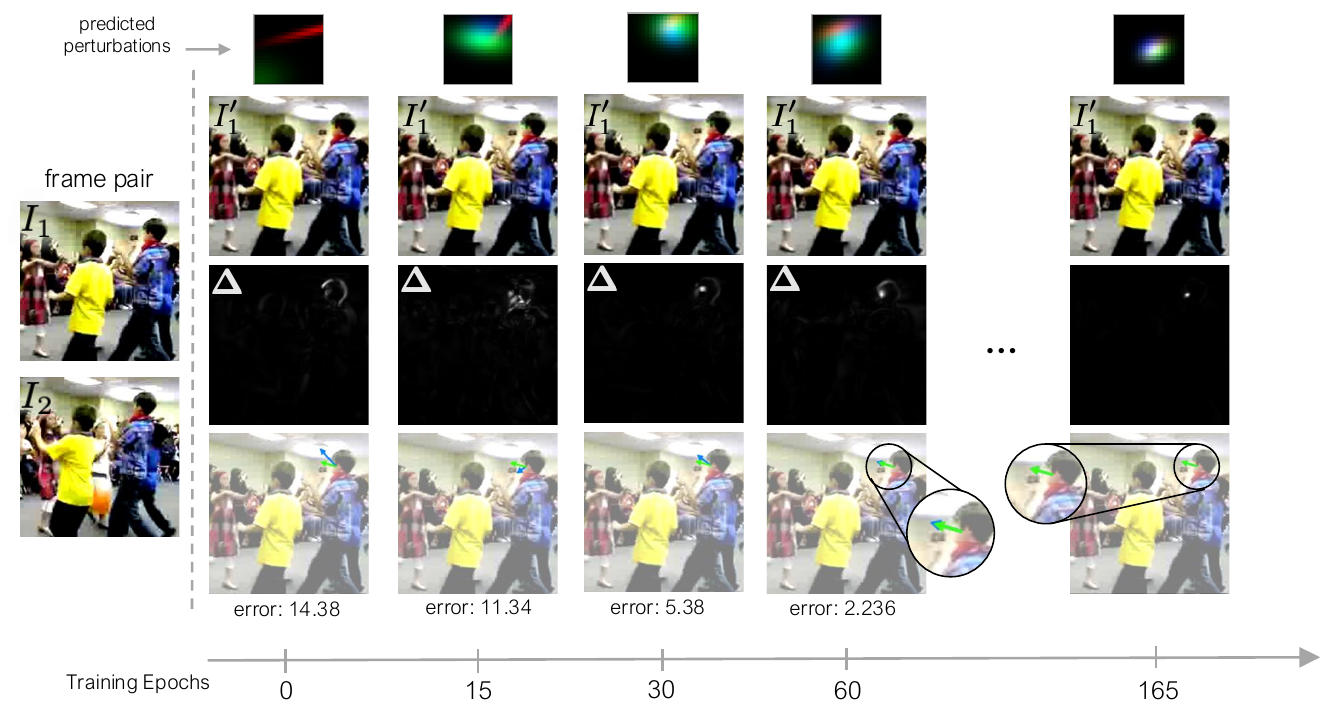}
    \caption{\textbf{Evolution of perturbations across training epochs:} We observe how the predicted perturbations change as the model trains. The perturbation starts as a disjoint streak of colors and converges to a localized peak. This in turn increasingly concentrates the difference image $\boldsymbol{\Delta}$ and leads to better flow prediction. Green is the ground truth flow obtained from the TAP-Vid dataset, and blue is our model's prediction.}
    \label{fig:evolution}
\end{figure}

The performance of $\textbf{\texttt{FLOW}}_\theta$ is greatly dependent on the quality of its learned Gaussian perturbations. In Figure~\ref{fig:evolution}, we see that the appearance of the perturbation evolves alongside the training of \ourmodel. As the perturbation converges into an optimal patch bespoke for the input frame, the quality of the flow prediction improves in tandem.

\end{document}